\documentclass[conference]{IEEEtran}
\IEEEoverridecommandlockouts
\usepackage{cite}
\usepackage{amsmath,amssymb,amsfonts,bm}
\usepackage{graphicx}
\usepackage{booktabs}
\usepackage{multirow}
\usepackage{textcomp}
\usepackage{xcolor}
\def\BibTeX{{\rm B\kern-.05em{\sc i\kern-.025em b}\kern-.08em
    T\kern-.1667em\lower.7ex\hbox{E}\kern-.125emX}}
\begin{document}

\title{SR-TTA: Spatial-Redundancy Test-Time Adaptation for Interference-Robust Respiration Sensing}

\author{\IEEEauthorblockN{Jingyuan Liu\IEEEauthorrefmark{1}\IEEEauthorrefmark{2},
Zheng Chang\IEEEauthorrefmark{1},
Haoqiu Xiong\IEEEauthorrefmark{2},
Zhuangzhuang Cui\IEEEauthorrefmark{2},
and Sofie Pollin\IEEEauthorrefmark{2}}
\IEEEauthorblockA{\IEEEauthorrefmark{1}School of Computer Science and Engineering, University of Electronic Science and Technology of China, Chengdu, China}
\IEEEauthorblockA{\IEEEauthorrefmark{2}WaveCoRe, Department of Electrical Engineering (ESAT), KU Leuven, 3001 Leuven, Belgium}
}

\maketitle

\begin{abstract}
Future 6G networks aim to expose sensing as a native service by reusing
communication infrastructure. We study respiration sensing on a cell-free
massive multiple-input multiple-output (MIMO) base station, where a $64$-antenna channel must be fused into a
breathing waveform. The state-of-the-art hand-crafted fusion is
near-optimal in benign conditions. It collapses, however, under strong
in-band motion interference, whose frequency falls inside the
respiration band. We
show that a learned complex-weight beamformer
recovers respiration by spatial nulling, and that the remaining gap to a
per-recording oracle can be closed at deployment by label-free test-time
adaptation. Crucially, we identify which label-free signal makes this
work. Frequency- and variance-based criteria cannot separate an in-band
interferer from breathing. Our spatial-redundancy test-time adaptation
(SR-TTA), which maximizes consistency across random antenna subsets
under an out-of-band spectral veto, preserves benign performance in our tests. The respiration-rate error drops from $5.8$ to $0.8$ breaths
per minute (bpm) under simulated in-band
interference, and the pipeline maps onto the Open Radio Access Network
(O-RAN) architecture as O-RAN distributed-unit (O-DU) range-gating, an
adaptation xApp, and a
calibration rApp. On real testbed recordings, a one-time cross-subject
calibration plus SR-TTA reduces failures from $47\%$ to $7\%$, drawing level
with the hand-crafted combiner using label-free test-time adaptation.
\end{abstract}


\section{Introduction}
\label{sec:intro}

Future 6G systems aim to provide environmental awareness as a native
network capability~\cite{isac}. Passive respiration sensing can reuse
cell-free massive multiple-input multiple-output (MIMO) channel state
information (CSI) without additional spectrum or hardware~\cite{bsbreath,ssir}.
Open RAN (O-RAN) offers a potential deployment substrate: its RAN
Intelligent Controllers (RICs) host learned xApps for radio-unit energy
saving~\cite{eexapp} and cell-free resource orchestration~\cite{demir}.

The sensing chain reduces the high-dimensional CSI to a one-dimensional
breathing waveform through a fusion front-end. State-of-the-art front-ends rely
on hand-crafted physical signal processing: a range transform, weighted antenna
combining (WAC) by breathing-to-noise ratio~\cite{bsbreath}, and a
frequency-separation rule based on the sensing-signal-to-interference ratio
(SSIR) to reject interference (SSIR-WAC)~\cite{ssir}. These
methods are
near-optimal for a single, static subject. Real deployments, however,
contain motion interference from a person turning over, walking slowly,
or a moving pet. Such co-occurring activity is a recognized failure mode in
wireless human sensing~\cite{fallattention}. When the motion is slow, its
frequency falls inside the respiration band ($0.1$--$0.5$\,Hz); when it
is also strong, both signal-to-noise weighting and frequency separation break
down, because the interferer is indistinguishable from breathing by power or
by frequency.

A natural response is to learn the fusion front-end. Deep models
operating directly on CSI have become the workhorse of wireless sensing, from
cross-domain gesture recognition~\cite{widar3} to CSI recovery and
prediction~\cite{csibert2}. Their known weakness is domain shift, as accuracy
degrades on unseen subjects, geometries, and environments~\cite{widar3}.
Test-time adaptation (TTA) offers a remedy by adapting the model at
deployment without labels~\cite{tent,ttt}, but its standard objectives
are designed for classification, not waveform recovery.

Our approach exploits the spatial diversity of distributed antennas.
BeamNet learns complex weights that can null interference through phase
cancellation, unlike nonnegative scalar fusion. Label-free test-time
adaptation refines these weights for each recording. The key question is
which adaptation criterion distinguishes breathing from an in-band
interferer: we use agreement across antenna subsets, rather than spectral
concentration or dominant variance.

The main contributions of this paper are summarized as follows:
\begin{itemize}
\item We study respiration sensing on a cell-free massive-MIMO
  base station with eight distributed access points (APs), where strong
  in-band motion interference defeats hand-crafted fusion. The sensing pipeline maps
  onto standard O-RAN components, with range-gating at the O-RAN distributed
  unit (O-DU), adaptation as a near-real-time (near-RT) RIC xApp, and one-time
  calibration as a non-real-time (non-RT) rApp.
\item We combine supervised BeamNet training with label-free SR-TTA. SR-TTA adapts the combiner per recording by maximizing
  consistency across random antenna subsets under an out-of-band spectral
  veto, without ground truth or an interferer detector, and we explain why
  frequency and subspace criteria cannot match it.
\item Experiments show that the proposed front-end recovers respiration where
  SSIR-WAC collapses (absolute Pearson correlation $|\rho|$ from $0.67$ to $0.95$, respiration-rate error
  from $5.8$ to $0.8$ bpm), repairs severe cross-domain shift on real
  recordings (failures $47\%\!\to\!7\%$), and improves with the number of
  distributed APs while the frequency baseline plateaus.
\end{itemize}

\section{System Model and Problem}
\label{sec:model}

\begin{figure}[t]
  \centering
  \includegraphics[width=0.9\columnwidth]{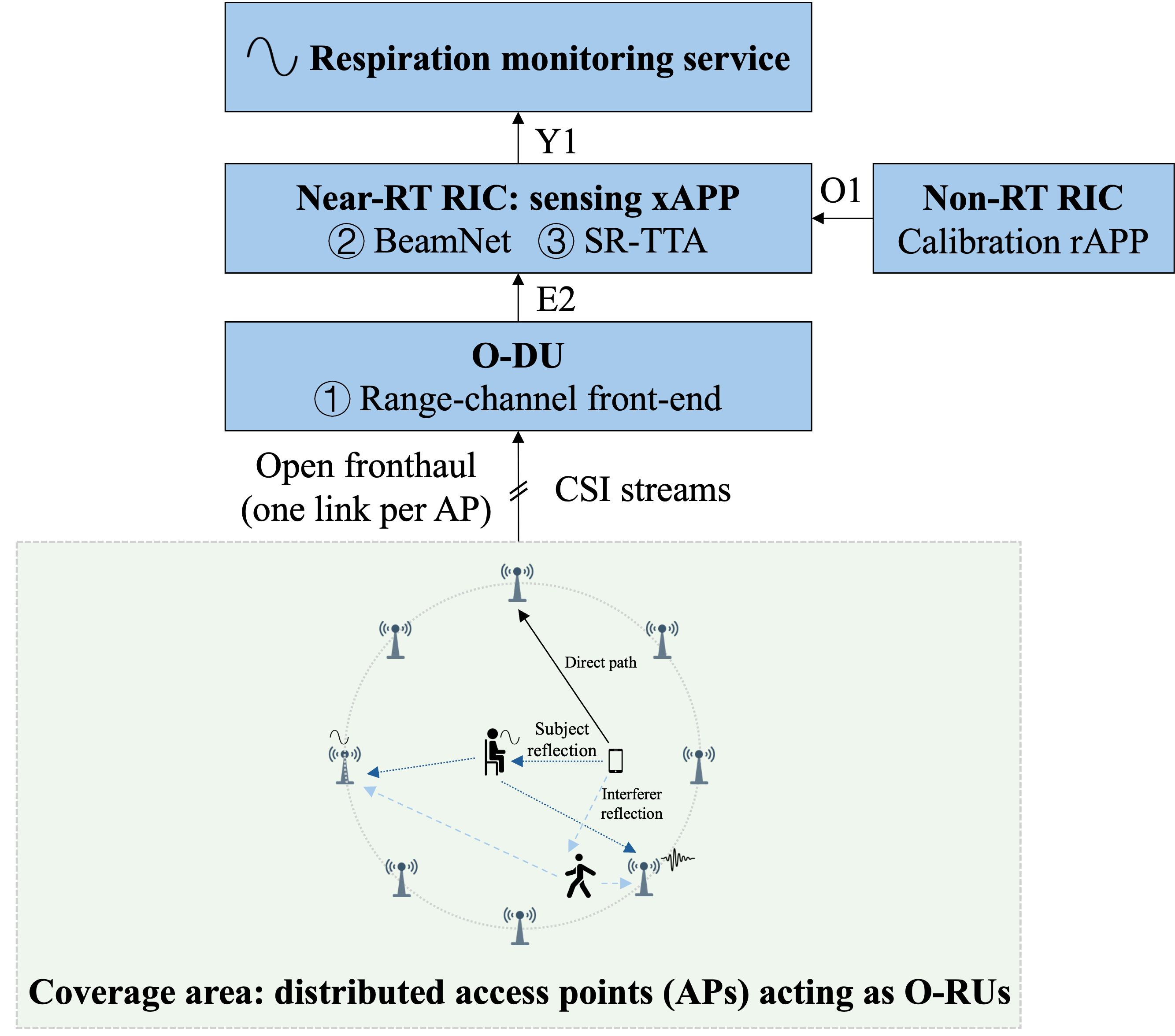}
  \caption{Considered scenario and proposed O-RAN embedding. \emph{Bottom:} eight
  distributed APs surround a seated subject, and a second moving body
  adds an in-band interferer. \emph{Top:} the pipeline mapped onto
  O-RAN, with range-gating \textbf{(1)} at the O-DU, BeamNet \textbf{(2)} and
  SR-TTA \textbf{(3)} as a near-RT RIC xApp, and one-time calibration as a
  non-RT rApp feeding the exposed respiration service.}
  \label{fig:system}
\end{figure}

Fig.~\ref{fig:system} summarizes the considered scenario. A cell-free
massive-MIMO base station with distributed APs (mapped to O-RAN radio units,
O-RUs) serves a user
equipment whose uplink pilots illuminate the scene. Each AP receives
the direct path, the breathing-modulated chest reflection of a seated subject,
and the reflection of a second moving body, which acts as the in-band
interferer formalized below.

\subsection{Cell-Free MIMO Uplink Sensing}
We reuse a communication cell-free massive-MIMO orthogonal
frequency-division multiplexing (OFDM) base station for passive
sensing~\cite{bsbreath}. A single-antenna user equipment transmits uplink
pilots, from which the base station estimates the CSI over the full band at
rate $f_{\mathrm{csi}}$, yielding the complex tensor
\begin{equation}
  \mathbf{H}\in\mathbb{C}^{M\times K\times I},
  \label{eq:csi}
\end{equation}
with $M$ distributed antennas grouped into uniform linear arrays at the
APs, $K$ subcarriers spanning a bandwidth $B$ at carrier $f_c$, over
$I$ frames. A subject seated near the user equipment
modulates the channel: the channel frequency response at antenna $m$, subcarrier
$k$, frame $i$ is a sum over $L$ propagation paths~\cite{bsbreath},
\begin{equation}
  h_{mk}[i]=\sum_{l=1}^{L}\alpha_{mkl}[i]\,
  e^{-j2\pi (f_c+k\Delta f)\,\tau_{ml}[i]},
  \label{eq:cfr}
\end{equation}
where for the chest path the delay carries the chest displacement $b[i]$,
$\tau_{ml}[i]=(\dot{d}_{ml}+\zeta_{ml}\,b[i])/c$, with $\dot{d}_{ml}$ the
static path length and $\zeta_{ml}$ a geometry projection factor. Breathing
thus appears as a slow, geometry-dependent phase rotation of the target
component.

\subsection{Range-Channel Extraction}
The path-length resolution $c/B\!\approx\!16.7$\,m is too coarse to separate
chest, direct, and interferer paths in this scene. An inverse discrete
Fourier transform (DFT) across subcarriers retains one range bin per antenna
for data reduction, not source separation. Static-component removal
(Sec.~\ref{sec:bf}) suppresses the static background. This produces one
complex range-channel per antenna and (downsampled) frame,
\begin{equation}
  r_m[t]\in\mathbb{C},\quad
  \mathbf{S}=[\,r_1;\dots;r_M\,]\in\mathbb{C}^{M\times T},
  \label{eq:range}
\end{equation}
$t\!=\!1,\dots,T$ at a decimated waveform rate $f_s\!\ll\!f_{\mathrm{csi}}$.
Breathing traces a small arc of a circle in the
complex plane of $r_m[t]$. Which real or imaginary projection is informative
varies per antenna with geometry~\cite{farsense}, and many antennas are
individually unreliable.

\subsection{In-Band Interference}
A second moving body, such as a person turning over, a person walking slowly,
or a pet, adds its own moving scatterer. Each range-channel then superposes
breathing, interferer, and noise,
\begin{equation}
  r_m[t]=r^{\mathrm{br}}_m[t]+r^{\mathrm{int}}_m[t]+n_m[t],\qquad
  r^{\mathrm{int}}_m[t]\approx c[t]\,e^{j\psi_m},
  \label{eq:superpos}
\end{equation}
where $c[t]$ is the interferer's shared temporal signature and $\psi_m$ its
antenna-dependent phase set by geometry. We call the interference in-band
when its motion spectrum overlaps the respiration band
$\mathcal{B}\!=\![0.1,0.5]$\,Hz.

\subsection{Problem Statement}
Given $\mathbf{H}$ (equivalently the range-channels $\{r_m[t]\}$), recover the
one-dimensional breathing waveform $b[t]$ up to sign and scale, robust to
in-band interference and without per-deployment ground truth.

\section{Interference-Robust Sensing Front-End}
\label{sec:method}

\begin{figure*}[t]
  \centering
  \includegraphics[width=0.8\textwidth]{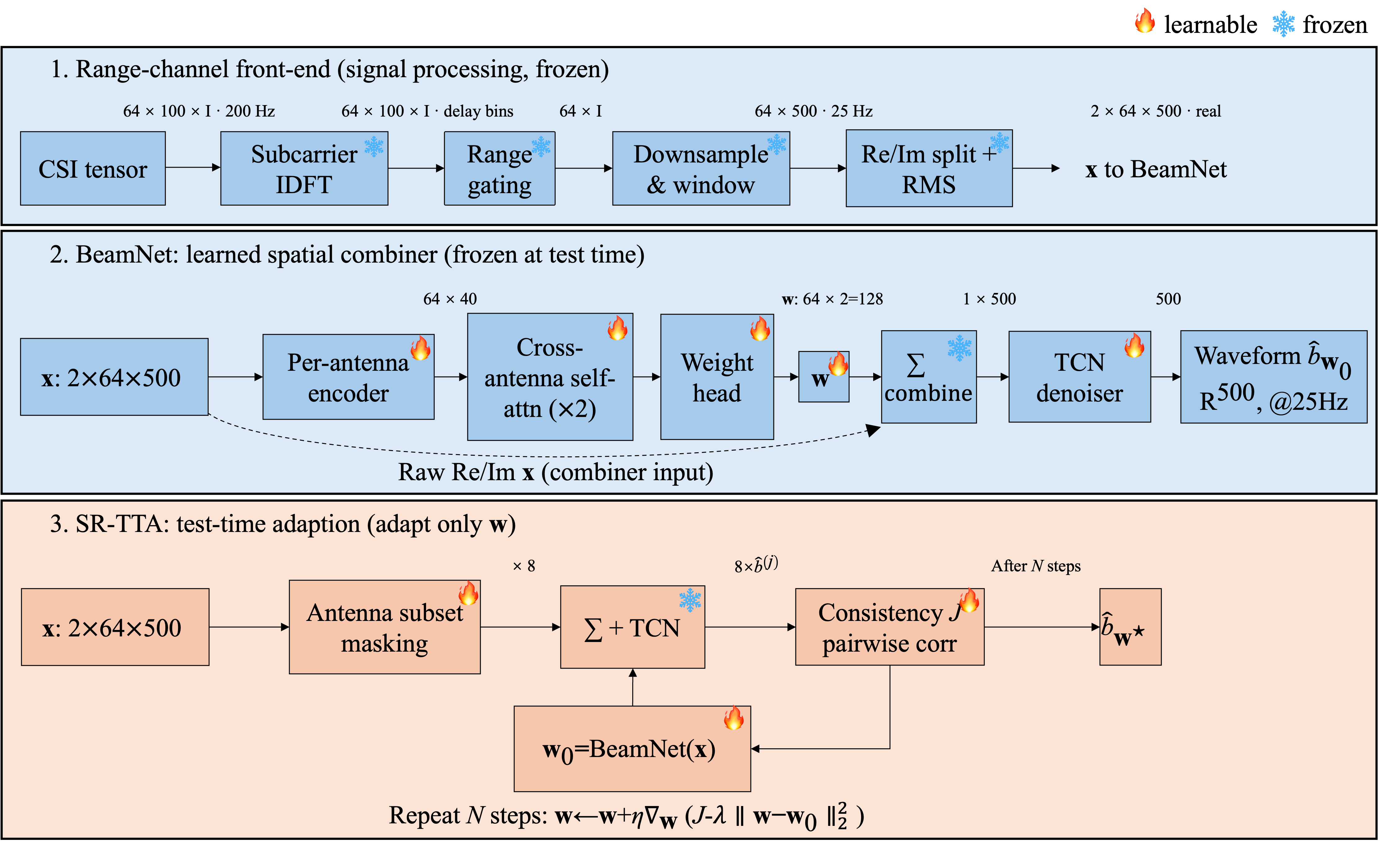}
  \caption{Proposed pipeline. \textbf{(1)} A frozen front-end reduces the raw
  CSI to the per-antenna range-channel
  $\mathbf{x}\!\in\!\mathbb{R}^{2\times M\times T}$. \textbf{(2)} BeamNet
  emits $M$ complex combiner weights ($2M$ reals) via per-antenna encoding
  and cross-antenna self-attention, and the combined trace is denoised by a
  temporal convolutional network (TCN) into the breathing waveform (trained on ground truth, frozen at test
  time). \textbf{(3)} SR-TTA adapts \emph{only} the $2M$ weights per
  recording, label-free, by maximizing agreement across random antenna
  subsets.}
  \label{fig:pipeline}
\end{figure*}

We treat respiration extraction as learning a per-recording spatial
combiner that maps the $M$-antenna range-channel to a single breathing
waveform (Fig.~\ref{fig:pipeline}). 

\subsection{Range-Channel and the Combiner Function Class}
\label{sec:bf}

We remove the static (line-of-sight) component of each $r_m$ and scale each
antenna to unit root-mean-square (RMS) amplitude, and write the real and imaginary parts as a
$2M$-dimensional feature per frame,
\begin{equation}
  \mathbf{f}[t] = \big[\,\mathbf{x}^{\mathrm{re}}[t]\,;\,\mathbf{x}^{\mathrm{im}}[t]\,\big]\in\mathbb{R}^{2M},
  \label{eq:feat}
\end{equation}
where $x^{\mathrm{re}}_m,x^{\mathrm{im}}_m$ are the normalized real/imaginary
parts of $r_m$. A combiner assigns each antenna a complex weight
$\mathbf{w}_m\!=\!(a_m,b_m)$, with $a_m,b_m$ its real and imaginary parts, and
applies it \emph{linearly},
\begin{equation}
  \hat{u}_{\mathbf{w}}[t]=\sum_{m=1}^{M}\big(a_m x^{\mathrm{re}}_m[t]+b_m x^{\mathrm{im}}_m[t]\big)
  =\mathbf{w}^{\!\top}\mathbf{f}[t],
  \label{eq:combine}
\end{equation}
where $\mathbf{w}=(\mathbf{a},\mathbf{b})$ collects the $2M$ weights
($\mathbf{a},\mathbf{b}\!\in\!\mathbb{R}^{M}$); these are exactly the
quantities that SR-TTA later adapts.
Equation~\eqref{eq:combine} is exactly the function class of the per-recording
supervised oracle, which recovers the waveform almost perfectly under
interference when fit to a recording's own ground truth. The
information needed for robust sensing is therefore present in $\mathbf{f}$ and
linearly recoverable, and the open question is how to find $\mathbf{w}$
without ground truth.

The complex weights are what separate this function class from scalar
combining. Under the interferer model~\eqref{eq:superpos}, a complex
combiner cancels the interferer whenever
\begin{equation}
  \sum_{m=1}^{M} w_m^{*}\,e^{j\psi_m}=0.
  \label{eq:null}
\end{equation}
This cross-antenna phase condition is unreachable by nonnegative
scalar weighting, which can down-weight a corrupted antenna but cannot
place a spatial null. It matters precisely under strong
in-band interference, where frequency-based fusion collapses.

\subsection{BeamNet: Learned Spatial Filtering}
\label{sec:beamnet}

BeamNet predicts the combiner from the input alone,
$\mathbf{w}_0=h_\theta(\mathbf{f})\in\mathbb{R}^{M\times 2}$. A shared
$1$D-convolutional encoder maps each antenna's normalized range-channel to a
descriptor $\mathbf{d}_m=\mathrm{Enc}_\theta(\mathbf{x}_m)\in\mathbb{R}^{D}$.
A learned positional embedding is then added, a self-attention encoder mixes
information \emph{across} the $M$ antennas, and a linear head emits
$\mathbf{w}_{0,m}$. The combined trace $\hat{u}_{\mathbf{w}_0}$
from~\eqref{eq:combine} is then
denoised by a temporal convolutional network (TCN) $g_\phi$ to yield the
breathing estimate $\hat{b}_{\mathbf{w}_0}=g_\phi(\hat{u}_{\mathbf{w}_0})$.
BeamNet is trained by maximizing waveform correlation,
\begin{equation}
  \min_{\theta,\phi}\;\;\mathbb{E}\!\left[\,1-\rho_{\mathrm{corr}}\big(\hat{b}_{\mathbf{w}_0},\,b^{\mathrm{GT}}\big)\right],
  \label{eq:trainloss}
\end{equation}
where $\rho_{\mathrm{corr}}(\cdot,\cdot)$ is the temporal correlation of
zero-mean, unit-norm signals and $b^{\mathrm{GT}}$ the ground-truth (GT)
waveform. BeamNet is pretrained on synthetic recordings
with a physics-based in-band interferer. When installed on a new site, it is
calibrated once, with the same loss, on clean recordings of
subjects other than those to be monitored, so the monitored subject and the
interference are never seen in training. This one-time deployment-domain
calibration anchors the learned prior to the deployment channel, and every
adaptation beyond it is label-free. A single feed-forward beam, however,
does not reach the per-recording oracle, because it must generalize to
unseen subjects and interferer geometries in one shot.

\subsection{Adaptation Criteria and Proposed SR-TTA}
\label{sec:tta}

At deployment there is no ground truth, yet the oracle gap is per-recording. We
therefore treat the combiner as a free per-recording variable, initialize it at
the network's beam $\mathbf{w}_0=h_\theta(\mathbf{f})$, freeze the TCN
$g_\phi$, and refine $\mathbf{w}$ for $N_{\mathrm{step}}$ steps to maximize a label-free
surrogate $J$ within a trust region around $\mathbf{w}_0$,
\begin{equation}
  \mathbf{w}^{\star}=\arg\max_{\mathbf{w}}\;\;
  J\big(\hat{b}_{\mathbf{w}}\big)\;-\;\lambda\,\lVert\mathbf{w}-\mathbf{w}_0\rVert_2^2 .
  \label{eq:tta}
\end{equation}
The choice of $J$ is the crux: it must distinguish breathing from an in-band
interferer without labels.

\subsubsection{Frequency Criteria Fail in Band}
A spectral band signal-to-noise-ratio (band-SNR) surrogate
$J_{\mathrm{freq}}=\log\!\big(\sum_{f\in\mathcal{B}}P_{\hat b}(f)\big/\sum_{f\notin\mathcal{B}}P_{\hat b}(f)\big)$,
or a within-band peakiness (spectral-entropy) surrogate, rewards energy
concentrated in the respiration band $\mathcal{B}$. But the in-band interferer
lives inside $\mathcal{B}$, so maximizing $J_{\mathrm{freq}}$ can lock
onto the interferer, the same structural failure as SSIR-WAC. Empirically these
surrogates degrade the beam. The failure is
specific to frequency used as an attractor, and frequency re-enters
safely as an out-of-band veto below.

\subsubsection{Subspace Nulling Fails under Overlap}
One can estimate the interferer subspace as the top-$k$ eigenvectors
$\mathbf{U}_k$ of the spatial covariance
$\mathbf{R}=\tfrac{1}{T}\mathbf{F}\mathbf{F}^{\!\top}$
($\mathbf{F}\!=\![\mathbf{f}[1],\dots,\mathbf{f}[T]]$) and null it,
$\mathbf{w}=(\mathbf{I}-\mathbf{U}_k\mathbf{U}_k^{\!\top})\mathbf{w}_0$. This
approach assumes that the dominant-variance directions belong to the
interferer alone. In our setting, however, the breathing and interferer
subspaces overlap, so the projection removes breathing energy together
with the interference, and performance degrades monotonically as $k$ grows.

\subsubsection{Proposed Spatial-Redundancy Consistency}
Neither spectrum nor dominant variance can thus separate an in-band interferer
from breathing. Our method, SR-TTA, uses a different,
discriminative signal, namely spatial redundancy. Breathing modulates
many of the $M$ distributed antennas, whereas a localized interferer
dominates only a few. A beam that ``rides'' the interferer therefore
relies on a small antenna set and is fragile to antenna dropout, while a
breathing beam is robust. We draw $S$ random keep-masks
$\mathbf{m}^{(j)}\!\sim\!\mathrm{Bernoulli}(p_{\mathrm{keep}})^{M}$ with keep probability $p_{\mathrm{keep}}$, form the
subset outputs
\begin{equation}
  \hat{b}^{(j)}_{\mathbf{w}}=g_\phi\!\Big(\textstyle\sum_{m}m^{(j)}_m\big(a_m x^{\mathrm{re}}_m+b_m x^{\mathrm{im}}_m\big)\Big),
  \label{eq:subset}
\end{equation}
and maximize their pairwise agreement,
\begin{equation}
  J_{\mathrm{cons}}(\mathbf{w})=\frac{2}{S(S-1)}\sum_{j<l}\rho_{\mathrm{corr}}\big(\hat{b}^{(j)}_{\mathbf{w}},\,\hat{b}^{(l)}_{\mathbf{w}}\big).
  \label{eq:cons}
\end{equation}
Maximizing~\eqref{eq:cons} pushes the beam onto the spatially-redundant
(breathing) component, information that neither the output spectrum nor the
dominant-variance subspace exposes.

\subsubsection{Out-of-Band Spectral Veto}
Spatial redundancy has one remaining blind spot, the mirror image of
SSIR-WAC's. Corruption seen coherently by all APs, such as the
subject's own non-respiratory motion or global channel fluctuation, is itself
spatially redundant, so $J_{\mathrm{cons}}$ cannot reject it. Such corruption
is, however, broadband or out-of-band in frequency. We therefore add a
spectral term only where it is trustworthy: a penalty on output energy outside
the service band $\mathcal{B}_{\mathrm{sv}}$ ($5$--$50$\,bpm). The adaptation
objective becomes
\begin{equation}
  J_{\mathrm{SF}}(\mathbf{w})=J_{\mathrm{cons}}(\mathbf{w})
  -\beta\,\Omega(\mathbf{w}),\qquad
  \Omega=\frac{\sum_{f\notin\mathcal{B}_{\mathrm{sv}}}P_{\hat{b}_{\mathbf{w}}}(f)}
  {\sum_{f}P_{\hat{b}_{\mathbf{w}}}(f)}.
  \label{eq:veto}
\end{equation}
The veto penalizes out-of-band energy without rewarding a particular
in-band peak. Spatial consistency distinguishes the sources within the band;
the veto suppresses broadband leakage. Neither term guarantees rejection
of an interferer that is both spatially redundant and in-band.

In summary, SR-TTA initializes $\mathbf{w}$ at the network beam
$\mathbf{w}_0$, runs $N_{\mathrm{step}}$ Adam steps on~\eqref{eq:tta} with
$J\!=\!J_{\mathrm{SF}}$, redrawing the $S$ masks at every step, and returns
the full-array output under the adapted weights. The procedure is label-free,
needs no interferer detector, and refines the beam within a trust region.
Pretraining and one-time calibration use labels; this test-time update does not.

\subsection{Complexity and O-RAN Placement}
\label{sec:oran}

We propose the following O-RAN mapping~\cite{demir}: range-channel
extraction at the O-DU, BeamNet and SR-TTA as a near-RT RIC xApp,
and offline calibration as a non-RT rApp. Retaining one complex sample
per antenna at $f_s$ reduces the stream by $Kf_{\mathrm{csi}}/f_s=800$
relative to the full subcarrier grid. Adaptation updates only $2M$ weights,
with computation scaling as $N_{\mathrm{step}}S$ TCN forward/backward
passes per window. The weights $\beta$ and $\lambda$ could be exposed as
policy parameters. This is an architectural proposal: O-RAN stack integration,
interface traffic, and end-to-end latency have not been benchmarked.
The $20$\,s observation window is distinct from adaptation runtime.

\section{Experiments}
\label{sec:results}

\subsection{Setup}
\label{sec:setup}

\subsubsection{Data}
We study a controlled in-band interference setting on the cell-free
massive-MIMO channel. A physics-based simulator renders each recording on the
distributed array model of the KU\,Leuven testbed: $M\!=\!64$ antennas in
eight APs (uniform linear arrays of eight), each $3$\,m from the
transmitter and spread in azimuth, with $K\!=\!100$ subcarriers over
$B\!=\!18$\,MHz at $f_c\!=\!3.51$\,GHz and CSI at
$f_{\mathrm{csi}}\!=\!200$\,Hz. It injects one moving scatterer (real multipath and
Doppler, not additive noise) with randomized reflectivity, position, and
amplitude ($0.8$--$2.0$\,m displacement), whose frequency is drawn inside the respiration band
($0.2$--$0.45$\,Hz). Each $40$\,s recording carries an independently drawn ground truth
(normal breathing or Kussmaul, $12$--$40$\,bpm), is decimated to
$f_s\!=\!25$\,Hz, and is cut into $20$\,s windows at a $10$\,s hop. Training and testing use
separately simulated recordings: $360$ for training ($1{,}080$ windows) and
$24$ disjoint recordings ($72$ windows) for reporting, with breathing,
geometry, and interferer parameters freshly drawn. For the mechanism study
we additionally sweep the interferer frequency from
out-of-band ($1.0$\,Hz) into the band ($0.2$\,Hz) with all $64$ antennas,
fixed $1$\,m displacement amplitude and reflectivity $5$ (12 recordings per
frequency). In contrast, Fig.~\ref{fig:scaling} and the strong-interference
row of Table~\ref{tab:regimes} use randomized in-band scenes (reflectivity $2$--$8$), so their averages
differ from this fixed-amplitude sweep. We also use $240$ clean synthetic
windows and $44$ real single-person recordings from the KU\,Leuven
testbed~\cite{bsbreath}, with $200$\,Hz CSI and $150$\,Hz motion-capture
(MoCap) respiration ground truth, resampled to $25$\,Hz.
Synthetic GT is the simulator's chest-displacement waveform.
The real+int. regimes inject
the same moving scatterer into the real recordings at a mild strength
($21$\,dB below the total channel power), in-band only or mixed. All
GT waveforms are band-passed to $5$--$50$\,bpm. All interfering motion
is simulated; the real recordings contain no second moving body.

\subsubsection{Metrics}
We report the absolute Pearson correlation $|\rho|$ to ground truth,
the failure rate (windows with $|\rho|\!<\!0.7$), and the
respiration-rate mean absolute error (MAE) in bpm. Rates are spectral peaks
of the estimate and GT on full recordings, using the same zero-padded
Kaiser-window periodogram (approximately $0.09$\,bpm grid). MAEs are rounded
to $0.1$\,bpm and describe agreement with GT-derived rates, not sensor
resolution. Figure bands are $\pm$one standard error of the mean (SEM).

\subsubsection{Baselines and Ceiling}
SSIR-WAC~\cite{bsbreath,ssir} is the training-free physics baseline, BeamNet
without adaptation is the feed-forward ablation, and SR-TTA is our
proposed method. The oracle is a ridge-regularized per-recording linear
combiner~\eqref{eq:combine} fit to a recording's own ground truth. It provides a supervised reference for linear fusion.

\subsubsection{Implementation}
BeamNet uses a width-$40$ per-antenna encoder, a two-layer self-attention mixer
over the $64$ antennas, and a dilated TCN ($\{1,2,4,8\}$), trained with
the correlation loss~\eqref{eq:trainloss} (AdamW). SR-TTA
uses $N_{\mathrm{step}}\!=\!40$ Adam steps (rate $0.05$), anchor $\lambda\!=\!0.3$, keep
probability $p_{\mathrm{keep}}\!=\!0.5$, $S\!=\!8$ antenna subsets, and veto weight
$\beta\!=\!0.5$ (any $\beta\!\in\![0.5,2]$ performs similarly). The same
configuration is used in every regime (no per-regime tuning, no interferer
detector). The deployment-domain calibration of Sec.~\ref{sec:beamnet} is
instantiated on the real testbed leave-one-group-out: the model evaluated on a
subject is fine-tuned ($50$ epochs, AdamW, $3\!\times\!10^{-4}$, noise and
antenna-dropout augmentation) only on other subjects' clean real
recordings and reused unchanged at every interference level. Synthetic
regimes use the pretrained model without calibration.

\subsection{Mechanism: Which Adaptation Criterion Survives In-Band?}
\label{sec:res-sweep}

\begin{figure}[t]
  \centering
  \includegraphics[width=0.8\columnwidth]{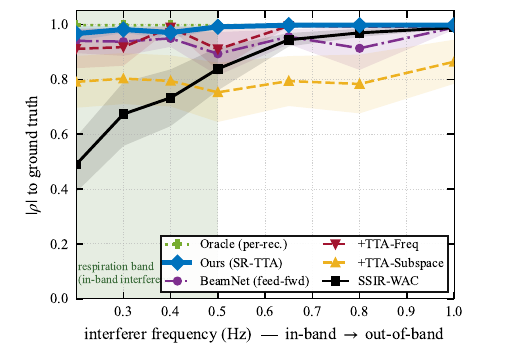}
  \caption{Adaptation criteria vs.\ interferer frequency (shaded = respiration
  band). SSIR-WAC collapses and frequency/subspace TTA fall short as the
  interferer enters the band, while only SR-TTA tracks the oracle throughout
  ($\pm$SEM bands).}
  \label{fig:sweep}
\end{figure}

\begin{figure}[t]
  \centering
  \includegraphics[width=0.8\columnwidth]{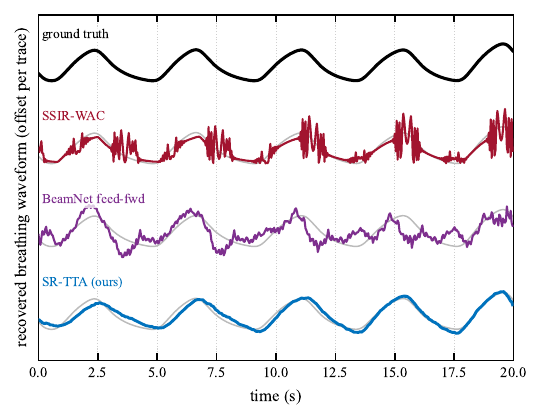}
  \caption{The in-band recording where adaptation helps most ($20$\,s, estimates
  offset over the ground truth): SSIR-WAC rides the interferer's bursts, the
  feed-forward beam drifts off breathing, and SR-TTA re-locks onto the true
  rhythm label-free (traces sign-aligned, unit-normalized).}
  \label{fig:waveform}
\end{figure}

Fig.~\ref{fig:sweep} sweeps the interferer across the band boundary.
Out of band ($\ge\!0.65$\,Hz) every method is near-perfect. The picture
inverts as the interferer enters the band. SSIR-WAC collapses
monotonically from $0.99$ to $0.49$ at $0.2$\,Hz, because its frequency-based
weighting cannot separate two signals that share the band. The two ``obvious''
label-free adaptations also fall short in band. Frequency TTA (band-SNR and
peakiness, T1/T2) cannot exceed the beam it starts from and stays below SR-TTA
($0.91$ vs.\ $0.97$ at $0.2$\,Hz). Dominant-variance subspace nulling (T3)
never helps ($0.75$--$0.86$), because the interferer and breathing subspaces
overlap. Only the spatial-redundancy criterion, SR-TTA, stays flat at
$0.97$--$1.00$ across the entire sweep, tracking the per-recording oracle. The
mechanism is spatial rather than spectral. An in-band interferer is still
spatially localized, so a beam that rides it is fragile to antenna dropout.
This is the signal that SR-TTA exploits and that the frequency and subspace
criteria cannot access.
Fig.~\ref{fig:waveform} makes this concrete on a single in-band recording.

\subsection{Spatial-Redundancy Scaling Law}
\label{sec:res-scaling}

\begin{figure}[t]
  \centering
  \includegraphics[width=0.8\columnwidth]{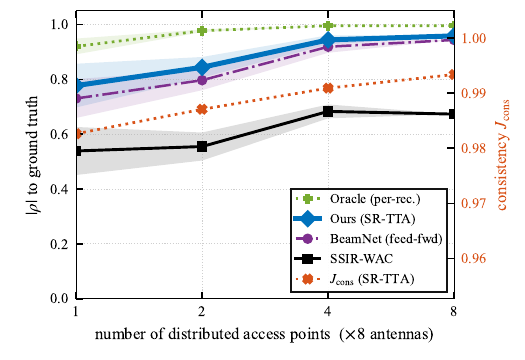}
  \caption{Scaling under in-band interference: $|\rho|$ vs.\ number of
  distributed APs (eight antennas each, $\pm$SEM over AP subsets),
  with the oracle ceiling and (right axis) SR-TTA's achieved consistency
  $J_{\mathrm{cons}}$. SSIR-WAC plateaus; SR-TTA scales.}
  \label{fig:scaling}
\end{figure}

If the mechanism is spatial redundancy, service quality should grow with the
number of distributed APs. Fig.~\ref{fig:scaling} repeats the strong in-band
experiment with $\#\mathrm{AP}\!\in\!\{1,2,4,8\}$ of eight antennas each
($M\!\in\!\{8,16,32,64\}$, averaged over AP subsets). SSIR-WAC barely
improves ($0.54\!\to\!0.67$), since scalar weighting cannot convert extra
antennas into the cross-antenna phase degrees of freedom needed for nulling.
SR-TTA instead climbs from $0.78$ to $0.96$, closely tracking the oracle
ceiling, which itself rises from $0.92$ to $1.00$: a larger array offers
more recoverable headroom, and it is the learned front-end that converts
this headroom into service quality. Throughout the sweep, the achieved
consistency $J_{\mathrm{cons}}$ stays near-saturated ($\geq\!0.98$),
confirming the premise that breathing is seen redundantly by the added APs.

\subsection{Cross-Domain Service Quality}
\label{sec:res-regime}

Finally, Table~\ref{tab:regimes} evaluates the same front-end across
regimes and connects fidelity to the exposed respiration-rate service. Under
strong in-band interference SR-TTA holds waveform fidelity where the physics
baseline breaks down ($0.95$ vs.\ $0.67$), and the respiration-rate error falls from
SSIR-WAC's $5.8$\,bpm to $0.8$\,bpm. Crucially,
the front-end does no harm off-design: on clean synthetic data all
front-ends stay near-perfect (failures $\le\!2\%$).

On the real testbed both methods are adapted to the deployment domain, so
the comparison is fair: SSIR-WAC was designed on this very
testbed~\cite{bsbreath,ssir}, BeamNet receives the one-time calibration of
Sec.~\ref{sec:beamnet}, and neither sees the test subject or labeled
interference. The synthetic-pretrained
beam reaches $|\rho|\!=\!0.50$ ($70\%$ failures), and calibration
lifts it to just $0.64$ ($47\%$). SR-TTA closes the remaining gap: at
$0.90$ and $7\%$ failures, it draws level with the hand-crafted combiner
($0.89$, $8\%$) on its home ground. Without calibration, SR-TTA reaches $0.89$ at $5$--$9\%$ failures,
suggesting that the veto mitigates part of the domain shift. In the milder
real+int.\ regimes, where the injected scatterer stays below the collapse
threshold, SSIR-WAC degrades (failures $8\%\!\to\!27\%$) and the adapted
front-end now edges it out on every metric (failures $24\%$ vs.\ $27\%$
and $14\%$ vs.\ $19\%$),
while the calibrated feed-forward beam alone still fails half its windows.
A single label-free front-end thus covers clean, interfered, and real
operation with no detector or baseline switching.

\begin{table}[t]
    \centering
    \caption{Cross-regime service quality: $|\rho|$ to GT, failure rate
    ($|\rho|\!<\!0.7$) per $20$\,s window, and respiration-rate MAE [bpm]; best
    in bold. Ours = BeamNet+SR-TTA; real+int. = real CSI with simulated interference.}
    \label{tab:regimes}
\footnotesize
\renewcommand{\arraystretch}{0.88}
\setlength{\tabcolsep}{3pt}
\begin{tabular}{llccc}
    \toprule
Regime & Metric & SSIR-WAC & BeamNet & Ours  \\\midrule
\multirow{3}{*}{in-band interf.\ (strong)} & $|\rho|$ & 0.67 & 0.94 & \textbf{0.95}  \\
 & fail$<$0.7 & 36\% & \textbf{4\%} & \textbf{4\%}\\
 & BPM MAE & 5.8 & \textbf{0.6} & 0.8 \\\midrule
\multirow{3}{*}{clean synthetic} & $|\rho|$ & 0.99 & 0.99 & \textbf{1.00}  \\
 & fail$<$0.7 & \textbf{0\%} & 2\% & \textbf{0\%}  \\
 & BPM MAE & \textbf{0.0} & 0.3 & \textbf{0.0} \\\midrule
\multirow{3}{*}{real (44 rec.)} & $|\rho|$ & 0.89 & 0.64 & \textbf{0.90}  \\
 & fail$<$0.7 & 8\% & 47\% & \textbf{7\%}  \\
 & BPM MAE & \textbf{0.1} & 1.8 & \textbf{0.1} \\\midrule
\multirow{3}{*}{real+int.\ (in-band)} & $|\rho|$ & 0.79 & 0.55 & \textbf{0.81}  \\
 & fail$<$0.7 & 27\% & 53\% & \textbf{24\%}  \\
 & BPM MAE & 0.7 & 2.7 & \textbf{0.6} \\\midrule
\multirow{3}{*}{real+int.\ (in+out)} & $|\rho|$ & 0.83 & 0.60 & \textbf{0.84}  \\
 & fail$<$0.7 & 19\% & 49\% & \textbf{14\%}  \\
 & BPM MAE & 0.6 & 2.9 & \textbf{0.5} \\\bottomrule
\end{tabular}
\end{table}

\section{Conclusion}
\label{sec:conclusion}

SR-TTA adapts a learned spatial combiner using label-free agreement across
antenna subsets and an out-of-band veto. Under simulated in-band interference,
respiration-rate MAE falls from $5.8$ to $0.8$\,bpm relative to SSIR-WAC.
After supervised calibration, adaptation reduces real-data failures from
$47\%$ to $7\%$, matching the hand-crafted baseline. Real moving-interferer
experiments and an implemented O-RAN control loop remain future work.

\section*{Acknowledgment}
This work is partly supported by the Fundamental and Interdisciplinary
Disciplines Breakthrough Plan of the Ministry of Education of China under
Grant No.~JYB2025XDXM116, and NSF of Sichuan under Grant No.~2026YFHZ0312.
This work is partly supported by the MultiX project under the European Union's
Horizon Europe research and innovation programme (Grant No.~101192521).
The work of Zhuangzhuang Cui is supported by the Research Foundation -- Flanders
(FWO), Senior Postdoctoral Fellowship under Grant No.~12AFN26N.



\begin{thebibliography}{00}
\bibitem{isac} F.~Liu \emph{et al.}, ``Integrated sensing and communications:
Toward dual-functional wireless networks for 6G and beyond,'' \emph{IEEE J.\ Sel.\
Areas Commun.}, vol.~40, no.~6, pp.~1728--1767, 2022.
\bibitem{bsbreath} H.~Xiong \emph{et al.},
``BS-Breath: Respiration sensing with cell-free massive MIMO,'' in \emph{Proc. IEEE ICASSP}, Hyderabad, India, 2025, pp.~1--5.
\bibitem{ssir} H.~Xiong \emph{et al.},
``Fundamentals and experiments of robust respiration sensing via cell-free
massive MIMO,'' \emph{IEEE J.\ Sel.\ Areas Commun.}, vol.~44, pp.~959--974,
2026.
\bibitem{eexapp} J.~Lu, P.~Yan, and H.~Zeng, ``EExApp: GNN-based reinforcement
learning for radio unit energy optimization in 5G O-RAN,'' in \emph{Proc.\
IEEE INFOCOM}, Tokyo, Japan, 2026, pp.~1--10.
\bibitem{demir} \"O.~T.~Demir \emph{et al.},
``Cell-free massive MIMO in O-RAN: Energy-aware joint orchestration of cloud,
fronthaul, and radio resources,'' \emph{IEEE J.\ Sel.\ Areas Commun.}, vol.~42,
no.~2, pp.~356--372, 2024.
\bibitem{fallattention} Y.~Xiao \emph{et al.}, ``Fall-attention: An
attention-based fall detection method for adjoint activities,'' \emph{IEEE
Trans.\ Mobile Comput.}, vol.~23, no.~7, pp.~7895--7909, 2024.
\bibitem{widar3} Y.~Zhang \emph{et al.}, ``Widar3.0: Zero-effort cross-domain
gesture recognition with Wi-Fi,'' \emph{IEEE Trans.\ Pattern Anal.\ Mach.\
Intell.}, vol.~44, no.~11, pp.~8671--8688, 2022.
\bibitem{csibert2} Z.~Zhao \emph{et al.}, ``CSI-BERT2: A BERT-inspired
framework for efficient CSI prediction and classification in wireless
communication and sensing,'' \emph{IEEE Trans.\ Mobile Comput.}, vol.~25,
no.~5, pp.~7241--7257, 2026.
\bibitem{tent} D.~Wang \emph{et al.},
``Tent: Fully test-time adaptation by entropy minimization,'' in \emph{Proc.\
ICLR}, 2021.
\bibitem{ttt} Y.~Sun \emph{et al.}, ``Test-time training with self-supervision for generalization under distribution shifts,'' in \emph{Proc.\ ICML}, vol.~119, 2020, pp.~9229--9248. 
\bibitem{farsense} Y.~Zeng \emph{et al.}, ``FarSense: Pushing the range limit of WiFi-based respiration sensing with CSI ratio of two antennas,'' \emph{Proc.\ ACM IMWUT}, vol.~3, no.~3, Art.~no.~121, 2019.
\end{thebibliography}
\end{document}